\documentclass[runningheads]{llncs}
\usepackage{float}
\usepackage{booktabs}
\usepackage{amsmath}
\usepackage[T1]{fontenc}
\usepackage{amssymb}
\usepackage{placeins}
\usepackage{xurl}
\usepackage{graphicx}
\usepackage{microtype}
\usepackage{hyperref} 
\begin{document}
\title{TRCGL-Net: A Long-Tailed Multi-Label Chest X-Ray Classification Framework with Generative Data Augmentation and Label Co-Occurrence Modeling}
\titlerunning{TRGCL-Net}
%
\author{
Tong Shao\inst{1}\textsuperscript{*}  \and
Hongshun Ling\inst{1}\textsuperscript{*} \and
Li Zhang\inst{1}\textsuperscript{†}  \and
Jinjing Wu\inst{1} \and
Junke Wang\inst{1} \and
Yuan Gao\inst{1}  \and
Fang Wang\inst{1} 
}

\institute{
School of Biomedical Engineering, South-Central Minzu University, Wuhan 430074, China\\
}
\authorrunning{Shao et al.}

\maketitle

\let\thefootnote\relax
\footnotetext{\textsuperscript{*}These authors contributed equally.}
\footnotetext{\textsuperscript{\dag}Corresponding author. E-mail: zhangli1996@163.com}

\begin{abstract}

\textbf{Objective:} Chest X-ray multi-label classification is a core task in intelligent medical imaging diagnosis. However, real clinical data often exhibit extreme long-tailed distributions, leading to degraded performance on rare diseases in tail classes. This issue is not only driven by data scarcity but also by two intrinsic factors:1) attenuation of tail-class lesion representations under complex anatomical backgrounds, and 2) dominance of head classes in modeling label co-occurrence relationships.

\textbf{Methods:} To address these challenges, we propose TRCGL-Net. First, a learnable text-guided conditional diffusion model is employed to generate high-quality tail-class chest X-ray image samples under disease semantic constraints, improving data diversity and realism of rare disease patterns while alleviating class imbalance and preserving pathology-consistent semantics.Second, a channel reweighting mechanism is introduced to perform feature recalibration by emphasizing disease-relevant feature channels, thereby improving feature discriminability under long-tailed distributions.A class-aware attention mechanism is further applied to generate class-specific attention maps, enabling the model to localize disease-relevant regions and focus on fine-grained lesion areas.Finally, a graph convolution network based on label co occurrence is introduced to establish an information propagation mechanism among categories.

\textbf{Results:} Experiments on the PadChest dataset show that the proposed method achieves a tail-class mAP of 0.4904, an overall mAP of 0.4408, and an mAUC of 0.8989, outperforming state-of-the-art methods.

\textbf{Conclusion:} TRCGL-Net effectively improves recognition performance for rare diseases under long-tailed distributions and mitigates the impact of extreme class imbalance in chest X-ray multi-label classification.

\textbf{Significance:} This framework improves performance under extreme class imbalance, particularly for rare disease classification.

\keywords{Chest X-ray   \and Long-tailed distribution \and Multi-label classification \and Conditional diffusion model \and Graph convolutional network \and Attention mechanism}
\end{abstract}
\section{Introducion}

Chest X-ray (CXR) images are widely used in clinical chest disease diagnosis due to their fast acquisition, low cost, and relatively low radiation exposure. However, similar to many medical imaging datasets, real-world CXR data exhibit a pronounced long-tailed label distribution, where a small number of common disease categories appear frequently in clinical reports, while most rare disease categories contain only limited positive samples~\cite{dong2026isbi,cxrlt2024survey,cxrlt2024noisy}. This imbalance poses significant challenges for deep learning methods, as models tend to be biased toward majority classes during training, while underrepresenting rare but clinically important tail classes. Therefore, CXR multi-label classification requires not only the simultaneous identification of multiple co-existing disease labels, but also robust discriminative capability under highly imbalanced class distributions, particularly for tail categories~\cite{ref4}.

In recent years, deep learning has significantly advanced the development of multi-label disease recognition and analysis in chest X-ray (CXR) imaging~\cite{ref5}. Convolutional Neural Networks (CNNs) are capable of extracting hierarchical visual features, such as pulmonary textures, mediastinal contours, and localized high-density lesions, thereby enabling the detection of common abnormalities~\cite{ref6}. Subsequently, several studies have improved CXR multi-label classification from a multi-scale feature extraction perspective. For instance, Yu et al.~\cite{ref7} enhanced model performance on lesions of varying sizes by integrating local lesion features with global thoracic structural information.Meanwhile, Transformer-based architectures leverage self-attention mechanisms to establish global dependencies across different image regions, allowing models to directly capture long-range anatomical structures and inter-lesion relationships. This effectively addresses the limitation of CNNs, which primarily rely on local convolution operations and gradually expand the receptive field, resulting in constrained global structural modeling capability~\cite{ref8}. In addition, to model latent dependencies among disease labels, Sun et al.~\cite{ref9} proposed a Label Correlation Transformer that learns inter-label relationships within chest diseases using a Transformer-based framework, thereby improving the reliability of multi-label prediction.However, most existing methods assume relatively balanced training data and insufficiently consider the long-tailed distribution that is prevalent in real-world clinical scenarios.

In real clinical datasets, tail classes are typically characterized by scarce samples, and lesions are often small and exhibit low contrast, making them susceptible to interference from ribs, soft tissues, and other background structures~\cite{ref10}. This leads to a gradual attenuation of effective feature representations for tail classes during deep feature extraction, making it difficult for models to learn stable and discriminative lesion features. Meanwhile, when label co-occurrence relationships are directly exploited for feature learning, noise embedded in low-level visual representations may be further amplified, thereby constraining the recognition performance of tail diseases. Existing methods such as re-sampling, loss re-weighting, and conventional data augmentation~\cite{ref11} can partially alleviate class imbalance; however, they remain insufficient in effectively enhancing disease-relevant features of tail classes and in fully capturing inter-label dependencies, ultimately limiting tail-class recognition performance.

Based on existing studies, long-tailed multi-label classification in chest X-ray (CXR) analysis still faces three major challenges:(1) Scarcity of tail-class samples leads to difficulty in learning stable disease representations.
In real clinical datasets such as PadChest, positive samples for certain tail diseases are extremely limited, making it difficult for models to learn stable and discriminative class-specific features. Although traditional re-sampling or data augmentation methods can increase sample size, they fail to incorporate disease-level semantic constraints. As a result, the generated samples may lose lesion-specific characteristics or introduce features inconsistent with real pathological patterns, thereby limiting effective representation learning for tail classes.(2) Insufficient tail-class feature representation and susceptibility to background interference.CXR images contain numerous prominent anatomical structures unrelated to disease, such as ribs, clavicles, cardiac silhouettes, and soft tissue shadows. These structures may overlap with lesion regions in both spatial location and intensity distribution, making it difficult for models to accurately extract tail-class features. Existing classification heads typically rely on global pooling or linear classification strategies, which are not adaptive to strengthening discriminative features across different disease categories. In particular, tail lesions are often characterized by subtle grayscale variations, boundary morphology, and fine-grained texture differences. These cues correspond to heterogeneous channel responses in deep feature maps. Relying solely on spatial aggregation of class features is insufficient to enhance channel-wise discriminative signals associated with tail diseases, thereby limiting recognition performance.(3) Insufficient exploitation of label co-occurrence in tail classes.
In multi-label settings, different disease labels may co-occur, forming structured dependencies that can potentially improve prediction calibration for tail classes~\cite{ref12}. However, existing methods for modeling label co-occurrence are typically designed around global label distributions and are thus dominated by frequent classes. Consequently, the learned label relationships tend to reflect co-occurrence patterns among common diseases, while tail classes fail to effectively benefit from informative associations with high-frequency labels. This ultimately limits consistent performance improvements for tail-class recognition.

To address these key challenges, we propose a TRCGL-Net framework for long-tailed multi-label classification of chest X-rays (CXR). First, we design a learnable text-guided conditional diffusion model that incorporates disease label semantics as generative constraints, guiding the model to synthesize tail-class samples that better match the true clinical distribution, thereby alleviating the severe data imbalance in rare categories. Second, a channel reweighting mechanism is introduced to recalibrate feature representations by emphasizing disease-relevant channels, thereby improving feature discriminability under long-tailed distributions. Moreover, a category-aware attention aggregation method was employed to generate class-specific spatial responses, enabling lesion localization and capturing subtle pathological patterns. Finally, a graph convolutional network based on label co-occurrence is incorporated to model label dependencies and calibrate prediction scores, improving prediction consistency for tail classes. The main contributions of this work are summarized as follows:

(1) To address the scarcity of tail-class samples in long-tailed chest X-ray (CXR) datasets, we propose a learnable text-guided conditional diffusion generative strategy. By imposing disease semantic constraints on the generation process of tail-class samples, the model improves the class relevance of synthesized images and produces samples that are better aligned with the real data distribution.

(2) To address the issues of weak lesion representations in tail classes and their susceptibility to interference from non-lesion regions, we enhance disease-relevant feature channels via channel reweighting and further improve tail-class discriminability through category-aware attention-based feature aggregation.

(3) To address the insufficient exploitation of label co-occurrence information in tail classes, we introduce a graph convolutional network based on label co-occurrence relationships. This module enables the propagation of discriminative information among correlated disease categories, allowing tail classes to leverage discriminative cues from frequently occurring related classes, thereby improving tail-class recognition performance.

\section{Related Work}

Long-tailed multi-label classification of chest X-ray (CXR) images has become a key task in intelligent medical image analysis, owing to its critical importance in disease screening, diagnosis, and clinical decision-making. The emergence of large publicly available datasets such as PadChes t~\cite{ref13}has significantly advanced the application of deep neural networks in this field, providing a solid foundation for automated lesion recognition and multi-label prediction. However, despite its large scale, the PadChest dataset still exhibits a pronounced long-tailed distribution, where a small number of frequent classes dominate most samples, while many rare classes contain only limited positive instances. Consequently, improving the recognition performance of tail classes under such imbalanced distributions remains a major challenge in CXR multi-label classification research.

In real clinical chest X-ray (CXR) data, samples of tail diseases are extremely limited, and their lesion regions often exhibit small size, low contrast, and blurred boundaries. As a result, the corresponding tail-class features are easily dominated by head-class representations during training and cannot be effectively preserved in the feature extraction process.To alleviate the performance degradation caused by data scarcity under long-tailed distributions, Sulake~\cite{ref14}proposed the LDAM-DRW loss to enhance the decision margins of tail classes, thereby improving tail-class recognition. Meanwhile, to address the limited availability of tail-class samples, Huang et al.~\cite{ref15}applied basic data augmentation techniques such as image flipping, cropping, and brightness adjustment to increase training data diversity. Furthermore, several studies have introduced generative models to synthesize additional minority-class chest X-ray samples. For instance, Kora and Ravula~\cite{ref16}employed DCGAN to generate synthetic tail-class samples, thereby mitigating class imbalance and improving the model’s ability to learn tail-class lesion representations.Despite these advances, most existing methods still focus on sample-level augmentation or loss reweighting strategies to alleviate class imbalance. However, they largely lack effective constraints from disease semantic information, which may lead to generated or augmented samples deviating from the key lesion characteristics of target diseases, thereby limiting the recognition performance for tail classes.

On the other hand, in chest X-ray (CXR) images, anatomical structures such as ribs, clavicles, cardiac silhouettes, and soft tissue shadows often overlap with lesion regions. For tail classes, abnormal regions are typically small in size and exhibit subtle intensity variations, making them more susceptible to interference from surrounding anatomical structures. Although conventional CNN-based methods can extract visual features, they often struggle to accurately distinguish tail-class abnormalities from adjacent normal structures.To enhance the model’s focus on abnormal regions, DualAttNet integrates image-level attention modules with fine-grained disease attention mechanisms to fuse global and local lesion-level classification information~\cite{ref17}. MBRANet introduces a residual architecture equipped with coordinate attention modules to extract multi-scale features, further combined with multi-scale feature fusion and multi-branch classifiers for chest disease classification~\cite{ref18}. CLARiTy generates class-specific attention maps to highlight abnormal regions and incorporates foreground segmentation with background suppression to reduce interference from complex anatomical structures~\cite{ref19}.Despite these advances in overall lesion recognition, existing methods still fail to sufficiently enhance disease-relevant feature representation in tail-class discrimination tasks.

Finally, in multi-label chest X-ray (CXR) image classification, there are typically complex co-occurrence relationships among disease labels; however, only a limited number of studies incorporate label co-occurrence knowledge into the model learning process. Chen et al. proposed a graph convolutional network-based label co-occurrence learning framework, CheXGCN~\cite{ref20}, which explicitly models inter-label co-occurrence relationships via a graph convolutional network and enhances label dependency learning through information propagation on the graph. Li et al. introduced the GL-MLL framework~\cite{ref21}, which leverages a label relation graph to model inter-class dependencies from both global static and local adaptive perspectives, thereby capturing label co-occurrence information more comprehensively. Cai et al. proposed a graph convolutional network with improved label semantics~\cite{ref22}, which integrates label semantic embeddings with graph structures to ensure that graph learning accounts not only for label co-occurrence but also for visual relevance and semantic consistency. Shi et al. proposed a graph-guided multi-scale cross-attention framework~\cite{ref23}, which combines a Vision Transformer (ViT) branch and a DenseNet-121 branch, and employs a multi-scale bidirectional dual cross-attention fusion module to align and interact between the two representations. In addition, a label graph is constructed based on conditional co-occurrence statistics from the training set, and GCN is used to refine label embeddings to model label dependencies.Although existing methods have achieved notable progress in general multi-label classification tasks, in long-tailed multi-label CXR classification where both class imbalance and label co-occurrence coexist, label co-occurrence information is still largely dominated by head-class relationships. This limits its effectiveness in improving tail-class recognition performance.

Overall, existing methods for long-tailed multi-label chest X-ray (CXR) classification still face several limitations, including the scarcity of samples for tail-class diseases, the small size of lesions that are easily confounded by background structures, and the insufficient exploitation of inter-disease relationships. To address these challenges, we propose a tail-aware long-tailed CXR multi-label classification framework that integrates a learnable text-guided conditional diffusion model, channel-enhanced and class-aware feature aggregation, and label co-occurrence–constrained discriminative optimization.

\section{Method}

\begin{figure}[htbp]
    \centering
    \includegraphics[width=0.85\textwidth]{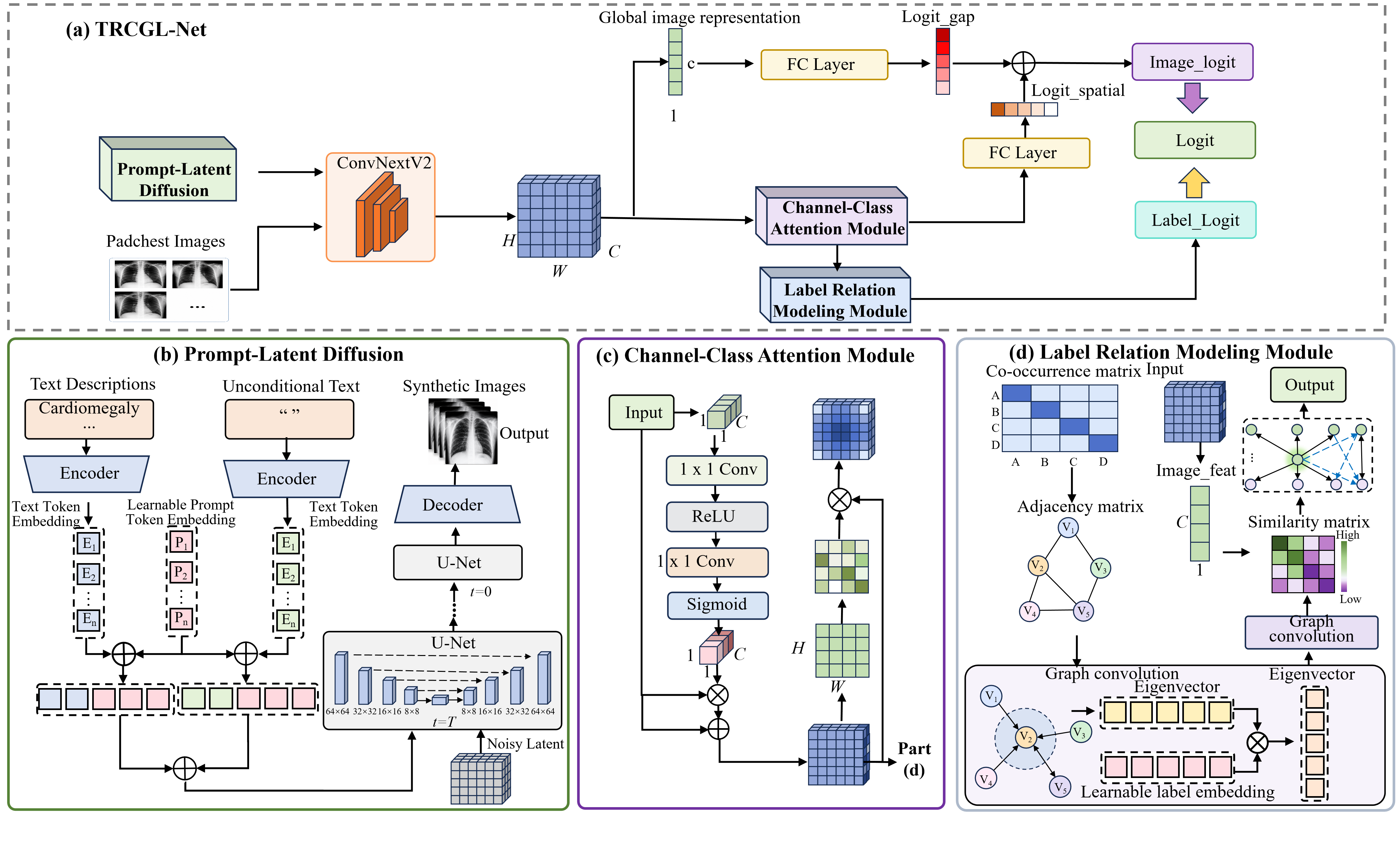}
    \caption{Overview of TRCGL-Net. (b)  Learnable text-guided prompt-latent diffusion with CLIP-based semantic conditioning and trainable context tokens for disease-guided tail-class synthesis under severe class imbalance. (c) ConvNeXtV2-based channel-class attention for enhancing tail-class discriminative features via channel reweighting and lesion-aware spatial . (d) GCN-based label co-occurrence modeling for inter-class dependency propagation and tail-class prediction refinement via exploiting label correlations. Final predictions are obtained via dual-level fusion}
    \label{fig:9}
\end{figure}

To enhance the recognition performance of tail-class diseases in long-tailed multi-label chest X-ray (CXR) classification, this study proposes the TRCGL-Net framework, as illustrated in Fig.~\ref{fig:9}. First, a learnable text-guided conditional diffusion generation module is employed to synthesize chest X-ray images for tail categories. Specifically, disease-category text prompts are combined with trainable prompt embeddings and fed into a text encoder to generate disease-aware semantic conditions, which guide a diffusion model to synthesize images exhibiting target pathological characteristics, thereby augmenting tail-class samples.Subsequently, both real and synthetic images are input into a ConvNeXtV2 backbone network for deep visual feature extraction.During the classification stage, channel reweighting and class-aware attention mechanisms are first applied to adaptively enhance discriminative feature representations across different categories, thereby strengthening representations for tail classes. On this basis, a label co-occurrence based class relationship graph is constructed. A graph convolutional network is then employed to propagate inter-class information among correlated disease categories, enabling tail classes to leverage discriminative cues from associated high-frequency classes. This effectively enhances the feature expressiveness of tail categories and improves their classification performance.

To improve training stability, a two-stage training strategy is adopted in this study. In the first stage, referred to as the classifier warm-up stage, the parameters of the ConvNeXtV2 backbone are frozen, and only the classification module and the label relationship modeling branch are trained. This allows the classification module to preferentially adapt to the multi-label classification task on the PadChest dataset while reducing disruptions to the pretrained backbone during the early training phase. In the second stage, a full fine-tuning strategy is applied, where all network parameters are unfrozen and jointly optimized, including the ConvNeXtV2 backbone, the channel-class attention module, and the label relation modeling module. This enables collaborative learning between visual feature extraction and label dependency modeling, thereby improving overall model performance.

\subsection{Prompt-Latent Diffusion}
To address the scarcity of tail-class samples in the PadChest dataset, this study proposes a learnable text-guided conditional diffusion generation strategy. Specifically, disease label texts are first semantically encoded, after which learnable context vectors are introduced to adaptively enhance the textual conditions. This design can be interpreted as a lightweight fine-tuning mechanism for the diffusion model. The resulting text embeddings are then used as conditional guidance for the diffusion model to generate images. In this way, high-quality training samples for tail classes can be synthesized while preserving the consistency of anatomical structures and lesion characteristics, thereby providing reliable tail-class sample support for downstream multi-label classification tasks.

For tail-class samples, disease-specific text prompts are first constructed based on the corresponding disease labels.In this study, tail classes are defined as categories with fewer than 500 training samples, reflecting the long-tailed nature of the dataset. Let the set of tail classes be denoted as \( y_{\text{tail}} = \{y_1, y_2, \dots, y_C\} \), where \( C \) represents the number of tail categories. For the \( n \)-th synthetic sample, its disease label is denoted as \( y_n \), and the corresponding text prompt is denoted as \( p_n \).

The text prompt, composed of disease names and radiology report descriptions from the PadChest dataset (e.g."Consolidation in the lower lung field consistent with pneumonia" or "No evidence of pleural effusion or cardiomegaly"). It is then tokenized using the CLIP tokenizer and encoded by the CLIP text encoder~\cite{ref24}, yielding a continuous vector representation $E_n$.

\begin{equation}
E_n = \text{CLIP}_{\text{TextEncoder}}(p_n), \quad E_n \in \mathbb{R}^{L \times d}
\end{equation}
where \( L \) denotes the fixed length of the text sequence and \( d \) denotes the dimensionality of the text embedding. This fixed-length formulation ensures compatibility with the cross-attention mechanism in diffusion models, where a constant-length key-value representation is required for stable attention computation. This process transforms the discrete label \( y_n \) into a continuous conditional representation, which can be directly used as textual conditioning input for the diffusion model.

Compared with traditional methods that rely on fixed text embeddings, this study introduces a set of learnable context vectors \( C = \{c_1, c_2, \dots, c_m\} \) to enhance tail-class feature representation, where \( m \) denotes the number of learnable context vectors and each vector shares the same dimensionality \( d \) as the text embeddings.

The learnable context vectors are concatenated with the text embeddings to form a conditionally augmented text sequence:
\begin{equation}
E_{cond} = \mathrm{Concat}_{seq}(C, E_n), \quad E_{\text{cond}} \in \mathbb{R}^{(m+L)\times d}
\end{equation}
where $\mathrm{Concat}_{\text{seq}}(\cdot)$ denotes concatenation along the sequence dimension. During training, the learnable context vectors are treated as optimizable parameters and jointly learned within the conditional diffusion framework.

Given a real image \( x_0 \), a forward diffusion process adds Gaussian noise \( \epsilon \) at timestep \( t \) to obtain a noisy image \( x_t \). The diffusion model then takes \( x_t \), the timestep \( t \), and the conditionally augmented text sequence \( E_{\text{cond}} \) as inputs to predict the added noise, yielding the following denoising objective:

\begin{equation}
\mathcal{L}_{\text{diff}} = \mathbb{E}_{x_0, t, \epsilon} \left[ \left\| \epsilon - \epsilon_\theta(x_t, t, E_{\text{cond}}) \right\|_2^2 \right]
\end{equation}
where $\|\cdot\|_2^2$ denotes the squared L2 (Euclidean) norm, i.e.,$\|x\|_2^2 = \sum_i x_i^2$, where $x_i$ is the $i$-th element of $x$. This corresponds to the standard mean squared error (MSE) objective. \( \epsilon_\theta(\cdot) \) denotes the noise prediction network of the diffusion model. The gradients of this loss are backpropagated to update the learnable context vectors \( C \), enabling them to progressively capture complementary tail-class information that is missing from the original text embeddings. This in turn improves the fidelity of generated images in terms of lesion morphology, intensity distribution, and local texture consistency with real tail-class samples.

During the image generation phase, the diffusion model is conditioned not only on the conditionally augmented text sequence to guide tail-class synthesis, but also on an unconditional text input to obtain a reference noise prediction that is independent of specific disease information. This unconditional branch helps stabilize generation and prevents over-reliance on a single text condition.

Subsequently, a guidance scale \( \omega \) is introduced to combine the two predictions:
\begin{equation}
\epsilon_{\text{cfg}} = \epsilon_{\text{uncond}} + \omega \left( \epsilon_{\text{cond}} - \epsilon_{\text{uncond}} \right)
\end{equation}
where \( \omega \) controls the influence of the textual condition during the generation process. A larger value of \( \omega \) strengthens the presence of target tail-class characteristics in the synthesized images, whereas a smaller value promotes greater diversity in intensity, texture, lesion location, and morphological appearance.
  
Given the combined noise prediction $\epsilon_{\text{cfg}}$, the reverse diffusion process proceeds by updating the latent representation at each timestep using the diffusion scheduler:
\begin{equation}
x_{t-1} = \sqrt{\bar{\alpha}_{t-1}} \hat{x}_0 + \sqrt{1 - \bar{\alpha}_{t-1}} \epsilon_{\text{cfg}}
\end{equation}
where $\hat{x}_0$ is the predicted clean latent reconstructed from $x_t$ and the predicted noise $\epsilon_{\text{cfg}}$, and $\bar{\alpha}_t$ denotes the cumulative noise scheduling coefficient at timestep $t$.

Finally, After $T$ denoising steps, the final latent $x_0$ is decoded by the pretrained VAE decoder to obtain the synthesized image:
\begin{equation}
I_{\text{syn}} = \mathrm{VAE}_{\text{Decoder}}(x_0)
\end{equation}

\subsection{Channel-Class Attention}

To alleviate the challenges of weak lesion representation, susceptibility to background interference in long-tailed multi-label chest X-ray (CXR) classification, this study enhance the deep features extracted by ConvNeXtV2 through a channel reweighting mechanism. This mechanism strengthens the activation responses of disease-relevant channels, thereby improving the discriminability of tail-class diseases.In addition, a class-aware spatial attention mechanism is introduced to generate independent attention maps for different disease categories. This allows each class to selectively focus on its corresponding lesion regions, enabling more effective extraction of localized pathological information from complex backgrounds and further improving recognition performance under long-tailed distribution settings.

Given an input chest X-ray image \( I \), a ConvNeXtV2 backbone network is first employed to extract high-level semantic features, yielding a feature map \( X \):
\begin{equation}
X = \phi(I; \theta), \quad X \in \mathbb{R}^{D \times H \times W}
\end{equation}
where \( \phi(\cdot; \theta) \) denotes the ConvNeXtV2 feature extraction network, \( \theta \) represents the model parameters, and \( D \), \( H \), and \( W \) denote the channel dimension, height, and width of the feature map, respectively. The backbone is initialized with pretrained weights and subsequently fine-tuned on the PadChest dataset.

Due to the presence of high-response anatomical structures in chest X-ray images, the model tends to prioritize head-class or background features under long-tailed class distributions, which may suppress lesion-specific representations of tail classes. To mitigate this issue, a channel-wise reweighting mechanism is applied to high-level semantic features extracted by the backbone, enhancing disease-relevant channel responses while preserving original information through residual connections.

Specifically, a global average pooling operation is applied to \( X \) to obtain a channel-wise global feature vector \( u \):
\begin{equation}
u = \frac{1}{HW} \sum_{i=1}^{H} \sum_{j=1}^{W} X_{i,j}
\end{equation}
where \( X_{i,j} \in \mathbb{R}^{D} \) denotes the feature vector at spatial location \( (i,j) \). A nonlinear transformation is then applied to generate channel attention weights \( g \):
\begin{equation}
g = \sigma \big( W_2 \delta (W_1 u) \big)
\end{equation}
where \( W_1 \) and \( W_2 \) are learnable parameters, \( \delta(\cdot) \) denotes the ReLU activation function, and \( \sigma(\cdot) \) denotes the sigmoid function. The channel weights \( g \) are used to modulate the response intensity of each channel for disease discrimination.

After obtaining the channel weights, a channel reweighting operation is applied to the backbone features, and the reweighted features are further fused with the original features via a residual connection \cite{ref25}.
\begin{equation}
X_r = X + X \odot g
\end{equation}
where \( \odot \) denotes element-wise multiplication and $g$ is broadcast along spatial dimensions.

Following channel reweighting, two parallel classification branches are constructed, including a global average pooling (GAP) branch and a class-aware spatial attention aggregation branch. In the GAP branch, global spatial pooling is applied to $X$ to obtain an image-level feature representation:
\begin{equation}
f_{\text{global}} = \frac{1}{|\Omega|} \sum_{(p,q)\in \Omega} X_{p,q}
\end{equation}
where \( \Omega = \{(p,q)\mid 1 \leq p \leq H, 1 \leq q \leq W\} \) denotes the spatial domain of the feature map. The resulting feature \( f_{\text{global}} \) is fed into a fully connected classifier to produce prediction scores \( s_{\text{gap}} \), leveraging global semantic information for classification.

To further capture localized discriminative regions for tail-class diseases, a class-aware spatial attention mechanism is introduced on the enhanced feature map \( X_r \). Specifically, a \( 1 \times 1 \) convolution is applied to obtain a class response map \( M \), where \( M_c^{p,q} \) denotes the response of class \( c \) at spatial location \( (p,q) \).

The spatial attention weights for each class are then computed via normalization:
\begin{equation}
a_c^{p,q} = \frac{\exp(M_c^{p,q})}{\sum_{(u,v)\in \Omega} \exp(M_c^{u,v})}
\end{equation}
where \( a_c^{p,q} \) measures the contribution of spatial location \( (p,q) \) to class \( c \) prediction. Based on these attention weights, class-specific local features are aggregated as:
\begin{equation}
v_c = \sum_{(p,q)\in \Omega} a_c^{p,q} X_r^{p,q}
\end{equation}

This mechanism enables different disease categories to focus adaptively on distinct spatial regions within the same chest X-ray, thereby enhancing lesion-specific representations and reducing interference from irrelevant anatomical structures.

The class-specific features \( v_c \) are then fed into a fully connected classifier to obtain spatial attention-based predictions \( s_{\text{spatial}} \). Finally, the image-level prediction is obtained by fusing the GAP and spatial attention branches:
\begin{equation}
s_{\text{img}} = s_{\text{gap}} + \lambda s_{\text{spatial}}
\end{equation}
where \( \lambda \) is a weighting coefficient controlling the contribution of the spatial attention branch. This fusion strategy integrates global semantic information and class-aware local lesion features, enabling the model to preserve overall anatomical context while enhancing sensitivity to subtle abnormalities in tail-class diseases.

\subsection{Label Relation Modeling}
In addition to image-based prediction, the model further exploits disease co-occurrence relationships to model label dependencies. Based on the same enhanced feature representation,  a label dependency modeling mechanism is introduced to capture statistical correlations among diseases. This design strengthens local lesion representations for tail classes while enabling inter-class information propagation through label co-occurrence, allowing tail classes to leverage additional discriminative cues from correlated high-frequency classes.

Let the multi-label annotation matrix of the training set be \( Y \in \mathbb{R}^{N \times C} \), where \( N \) denotes the number of samples and \( C \) denotes the number of disease categories. The label co-occurrence matrix is first computed as:
\begin{equation}
Q = Y^{\top}Y
\end{equation}
where \( Q_{ij} \) represents the frequency of co-occurrence between class \( i \) and class \( j \) within the training set.

To construct a robust and noise-resistant label graph, the raw co-occurrence matrix is first normalized to mitigate the bias introduced by highly frequent classes:
\begin{equation}
\tilde{Q}_{ij} = \frac{Q_{ij}}{\sum_{j} Q_{ij}}
\end{equation}

Then, a symmetrization operation is applied to ensure bidirectional consistency of label dependencies:
\begin{equation}
\tilde{Q} \leftarrow \frac{\tilde{Q} + \tilde{Q}^{\top}}{2}
\end{equation}

To further suppress spurious correlations, a thresholding operation is introduced:
\begin{equation}
\tilde{Q}_{ij} =
\begin{cases}
\tilde{Q}_{ij}, & \tilde{Q}_{ij} \ge \tau \\
0, & \tilde{Q}_{ij} < \tau
\end{cases}
\end{equation}

Self-loop connections are then added to preserve individual label semantics:
\begin{equation}
\tilde{Q} \leftarrow \tilde{Q} + I
\end{equation}

Finally, the label adjacency matrix is obtained via row-wise normalization:
\begin{equation}
A = D^{-\frac{1}{2}} \tilde{Q} D^{-\frac{1}{2}}, \quad
D_{ii} = \sum_{j} \tilde{Q}_{ij}
\end{equation}
where \( A \in \mathbb{R}^{C \times C} \) encodes the structured co-occurrence relationships among disease categories and serves as the message-passing graph for label graph convolution.

In the label graph convolution branch, each disease category is treated as a node in the graph and initialized with a learnable label embedding \( E_0 \in \mathbb{R}^{C \times d} \), where \( d \) denotes the embedding dimension. Two-layer graph convolution is performed over the label adjacency matrix \( A \) to propagate label dependencies, yielding class prototypes enriched with co-occurrence information:
\begin{equation}
E_1 = \delta \big( A E_0 W_0 \big)
\end{equation}

\begin{equation}
E_2 = A E_1 W_1
\end{equation}
where \( W_0 \) and \( W_1 \) are learnable parameters, and \( \delta(\cdot) \) denotes the ReLU activation function. The resulting \( E_2 \in \mathbb{R}^{C \times D} \) represents class prototypes updated with label co-occurrence relationships. Through this process, disease categories with strong co-occurrence relationships are able to exchange information in the label space, enabling tail classes to benefit from correlated high-frequency classes.

To obtain label-graph-based predictions, global average pooling is first applied to the enhanced feature map $X_r$, yielding an image-level feature representation:
\begin{equation}
g_r = \frac{1}{|\Omega|} \sum_{(p,q)\in \Omega} X_r^{p,q}
\end{equation}
where \( \Omega \) denotes the spatial domain of the feature map. The image-level feature \( g_r \) is then projected into the same embedding space as the class prototypes and matched with \( E_2 \), yielding label-graph-based prediction scores:
\begin{equation}
s_{\text{label}} = g_r E_2^{\top} + b_g
\end{equation}
where \( b_g \) is a bias term and \( s_{\text{label}} \in \mathbb{R}^{C} \) denotes the prediction vector from the label graph branch.

Finally, image-based predictions and label-graph-based predictions are combined via residual aggregation:
\begin{equation}
s = s_{\text{img}} + \alpha s_{\text{label}}
\end{equation}
where \( \alpha \) is a fusion coefficient controlling the contribution of the label graph branch. The term \( s_{\text{img}} \) provides global semantic and local lesion-aware visual information, while \( s_{\text{label}} \) incorporates disease co-occurrence dependencies. This fusion strategy enables the model to enhance discriminative capability from visual features while leveraging label correlations to supplement tail-class recognition, thereby improving long-tailed multi-label chest X-ray classification performance.

\section{Experiment}
\subsection{Dataset}
\subsubsection{Padchest Dataset}
The PadChest dataset is a publicly available chest X-ray (CXR) dataset released by San Juan Hospital, Spain. It contains imaging data from over 67,000 patients and more than 160,000 radiographic images, covering six different projection views. The dataset is annotated with 174 classes of radiological findings and 104 anatomical labels.

\begin{figure}[htbp]
    \centering
    \includegraphics[width=0.7\textwidth]{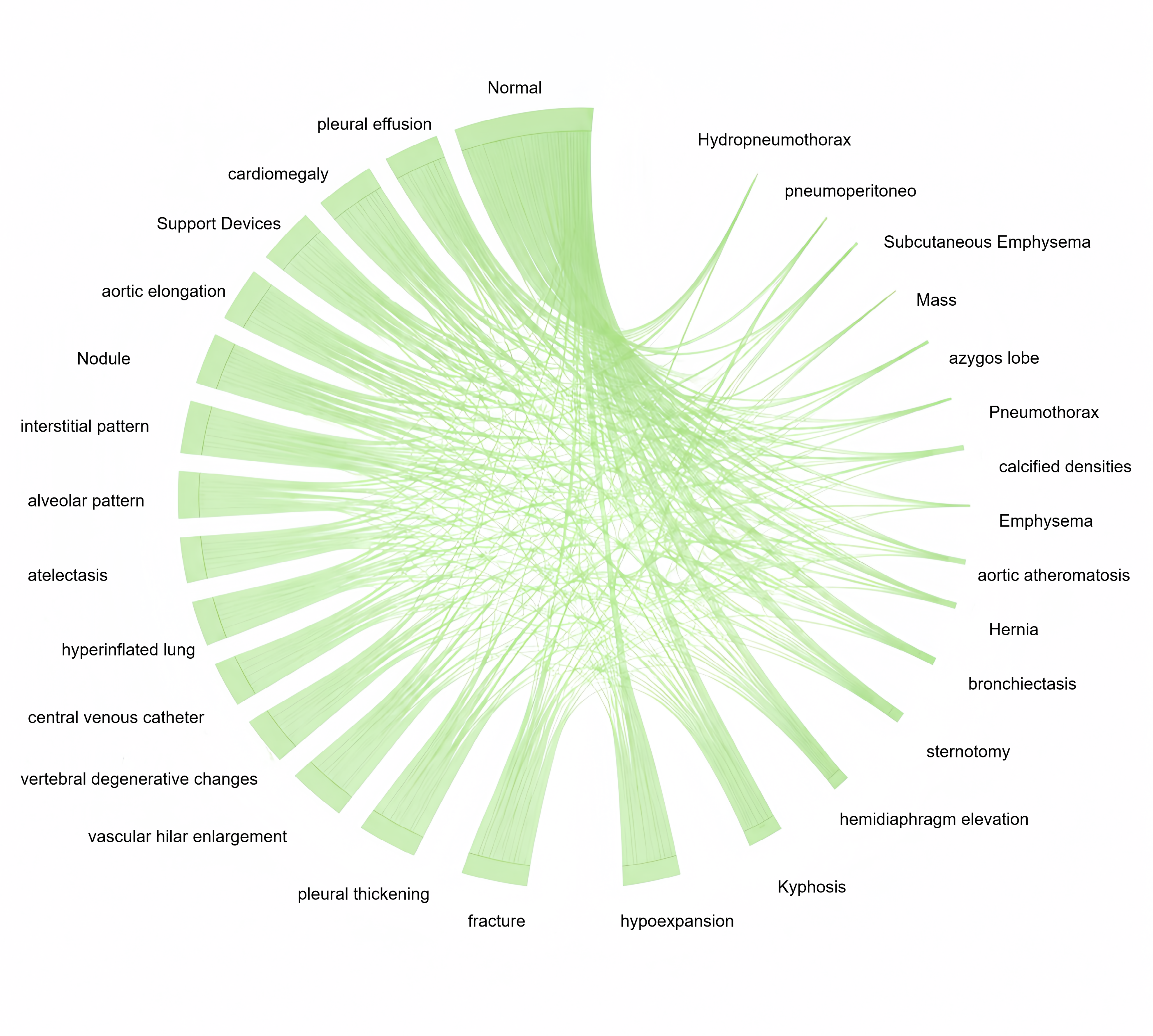}
    \caption{Label Distribution and Co-occurrence Analysis of 30 Classes in the PadChest Dataset}
    \label{fig:Padchest}
\end{figure}

To ensure a reasonable experimental split and data independence, the dataset was divided at the patient level into a training subset (80\%) and a validation subset (20\%). Fig.~\ref{fig:Padchest} illustrates the distribution of 30 chest X-ray labels in the PadChest dataset, as well as their co-occurrence relationships. In the chord diagram, each node on the circumference represents a distinct chest X-ray label, and the size of each corresponding sector reflects the number of samples in that category. The connections between nodes indicate the co-occurrence of two labels within the same radiograph; a denser set of links implies more frequent co-occurrence patterns between different disease labels.The analysis reveals a pronounced imbalance in the number of samples across different disease categories, exhibiting an overall long-tailed distribution. In addition, complex co-occurrence relationships are observed among certain labels, indicating that chest X-ray multi-label classification not only requires addressing the class imbalance problem but also necessitates explicitly modeling the dependencies among disease labels.

\subsubsection{Synthetic Image}
\begin{figure}[htbp]
    \centering
    \includegraphics[width=0.6\textwidth]{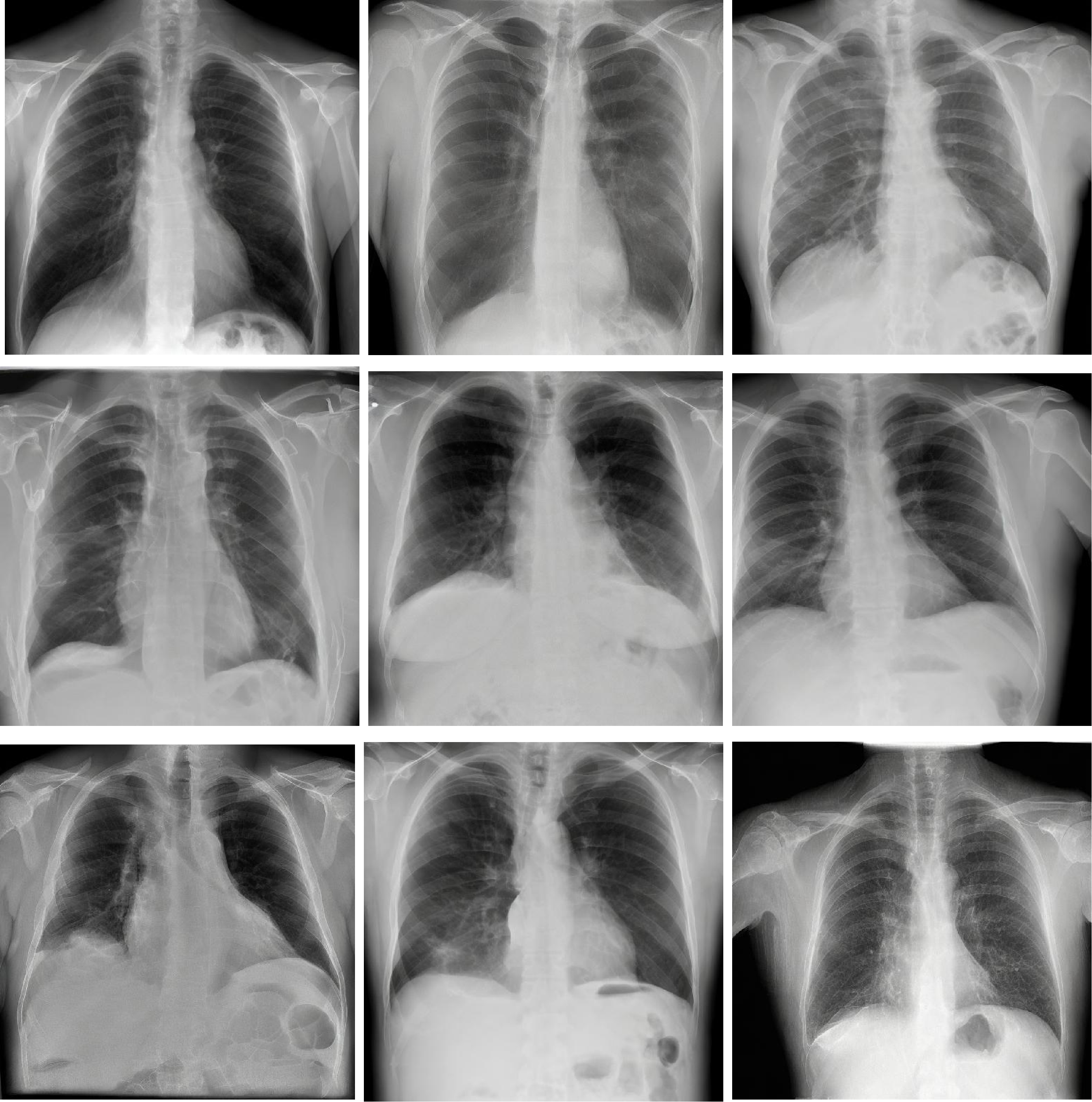}
    \caption{Synthetic chest X-ray images generated by a learnable text-guided diffusion model}
    \label{fig:synthetic}
\end{figure}

To address the scarcity of samples in the tail classes of the PadChest dataset, this study employs a text-guided conditional diffusion model to generate synthetic chest X-ray images for underrepresented categories. The resolution of the synthesized images is set to 512×512 pixels to ensure that anatomical structures and lesion characteristics remain consistent with those of real radiographs.Representative synthetic images are shown in Fig.~\ref{fig:synthetic}. The filtered synthetic images are then combined with the original training subset to form an augmented dataset for tail-class enhancement, which is subsequently used for multi-label classification training.
\subsection{Experimental setup}
All experiments were conducted on a computing platform equipped with an NVIDIA GeForce RTX 4090 GPU, and both model training and inference were implemented using the PyTorch framework. All input images were uniformly resized to 224 × 224 pixels and normalized to ensure training stability and reproducibility.

During the training process, a two-stage training strategy was adopted to improve training stability. The first stage was a classifier head warm-up phase, with 10 training epochs and a learning rate of $1 \times 10^{-3}$. The second stage was an overall fine-tuning phase with 50 training epochs. In this stage, the learning rate of the backbone network was set to $1.5 \times 10^{-5}$, while the learning rates for the classification head and the label-graph branch were set to $1.5 \times 10^{-4}$.

\subsection{Evaluation Metrics}

In long-tailed multi-label classification tasks, a single metric is insufficient to comprehensively reflect model performance. Therefore, this study adopts multiple evaluation metrics, including mean Average Precision (mAP), mean Area Under the ROC Curve (mAUC), and macro-averaged F1 score (macro-F1).

The macro-F1 score is defined as:
\begin{equation}
\text{Precision} = \frac{TP}{TP + FP}
\end{equation}

\begin{equation}
\text{Recall} = \frac{TP}{TP + FN}
\end{equation}

\begin{equation}
F1 = \frac{2 \times \text{Precision} \times \text{Recall}}{\text{Precision} + \text{Recall}} = \frac{2TP}{2TP + FP + FN}
\end{equation}

where True Positive (TP) denotes the number of samples correctly predicted as positive, False Positive (FP) denotes the number of negative samples incorrectly predicted as positive, and False Negative (FN) denotes the number of positive samples incorrectly predicted as negative. To measure the balance of model performance across all categories, the macro-F1 score is computed as:
\begin{equation}
\text{macro-F1} = \frac{F1_1 + F1_2 + \cdots + F1_C}{C}
\end{equation}

\textbf{mAP:}

Average Precision (AP) is used to evaluate the performance of each class across different thresholds of the precision-recall curve:
\begin{equation}
AP_c = (R_1 - R_0)P_1 + (R_2 - R_1)P_2 + \cdots + (R_n - R_{n-1})P_n
\end{equation}

where $P_n$ and $R_n$ denote the precision and recall at the $n$-th threshold, respectively, and $AP_c$ represents the average precision of class $c$. The mAP is defined as the mean of AP across all classes:
\begin{equation}
mAP = \frac{AP_1 + AP_2 + \cdots + AP_C}{C}
\end{equation}

\textbf{mAUC:}

The Receiver Operating Characteristic (ROC) curve and Area Under the Curve (AUC) are used to evaluate the model’s ability to distinguish between positive and negative samples:
\begin{equation}
TPR = \frac{TP}{TP + FN}
\end{equation}

\begin{equation}
FPR = \frac{FP}{FP + TN}
\end{equation}

where True Negative (TN) denotes the number of samples correctly predicted as negative. AUC represents the area under the ROC curve; values closer to 1 indicate stronger discriminative ability:
\begin{equation}
AUC = \frac{FPR_{i+1} - FPR_i}{2} \times (TPR_{i+1} + TPR_i)
\end{equation}

The mAUC is defined as the average AUC over all classes:
\begin{equation}
mAUC = \frac{AUC_1 + AUC_2 + \cdots + AUC_C}{C}
\end{equation}

Through these metrics, the overall performance and tail-class recognition capability of the proposed model in long-tailed multi-label chest X-ray classification can be comprehensively evaluated.

\subsection{Main Experiments}
To comprehensively evaluate the effectiveness of the proposed TRCGL-Net for long-tailed multi-label chest X-ray classification, we conduct systematic comparisons on the PadChest dataset against a range of mainstream deep learning baseline models as well as state-of-the-art algorithms in this domain. The compared methods include the classical convolutional architecture ResNet-50~\cite{ref25} , EfficientNet-B0~\cite{ref27} , modern convolutional and self-attention-based networks such as ConvNeXt~\cite{ref28}, ViT-B~\cite{ref29}, CLIP~\cite{ref30}, as well as advanced methods designed for medical long-tailed and zero-shot classification, such as ConvNeXt+LDAM~\cite{ref14}and ConvNeXt+CSRA~\cite{ref31}.

As shown in Table \ref{tab:padchest_results}, among the backbone networks without any dedicated long-tailed learning strategies, ConvNeXt demonstrates the strongest baseline discriminative capability, achieving significantly higher mAP and AUC than the traditional ResNet and EfficientNet. This indicates that modern pure convolutional architectures, by introducing large-kernel designs, effectively expand the receptive field and exhibit stronger multi-scale feature extraction capability when dealing with chest X-ray images characterized by large lesion scale variations and blurred boundaries.However, Table~\ref{tab:map_auc_comparison} further reveals a clear performance drop on tail classes, suggesting that standard models are biased toward head classes under severe data imbalance.

Although ViT and CLIP benefit from global attention and large-scale vision-language pretraining, their performance remains limited in capturing subtle and low-contrast lesions in tail classes when applied to long-tailed medical multi-label classification. Methods such as ConvNeXt+LDAM partially improve tail-class performance by rebalancing decision boundaries, but they fail to address the underlying issue of insufficient tail-class feature representation.When incorporating long-tailed learning strategies, ConvNeXt+LDAM, which incorporates a class-balanced loss strategy, effectively improves the overall F1-score by enhancing decision boundaries for tail classes. However, since this approach only imposes post-hoc constraints at the loss level, it does not address the insufficient feature representation caused by limited tail-class samples, and therefore its mAP is even slightly lower than that of the baseline ConvNeXt. In comparison, ConvNeXt+CSRA further strengthens performance through class-specific attention recalibration, achieving consistent gains across all metrics . This demonstrates that feature-level adaptive refinement is more effective than purely loss-level reweighting in mitigating long-tailed imbalance.

\begin{table}[htbp]
\centering
\caption{Performance comparison of different methods on the PadChest dataset}
\label{tab:padchest_results}
\begin{tabular}{lccc}
\hline
Method & mAP $\uparrow$ & mAUC $\uparrow$ & F1 $\uparrow$ \\
\hline
ConvNeXt~\cite{ref28}  & 0.3323 & 0.8720 & 0.2166 \\
ResNet~\cite{ref25}  & 0.2092 & 0.8371 & 0.0991 \\
EfficientNet~\cite{ref27}  & 0.1807 & 0.7717 & 0.0322 \\
CLIP~\cite{ref30}  & 0.2525 & 0.8291 & 0.0501 \\
ViT~\cite{ref29}  & 0.1901 & 0.7778 & 0.0376 \\
ConvNeXt + LDAM~\cite{ref14}  & 0.2962 & 0.8464 & 0.2752 \\
ConvNeXt + CSRA~\cite{ref31}  & 0.3671 & 0.8581 & 0.3227 \\
Ours & \textbf{0.4408} & \textbf{0.8989} & \textbf{0.4369} \\
\hline
\end{tabular}
\end{table}

\subsection{Quantitative Analysis}
To comprehensively evaluate the model’s robustness under different levels of data imbalance, disease categories in the PadChest dataset are stratified into three groups—head, medium, and tail—according to their training sample frequencies computed from the PadChest training set.Specifically, categories with more than 5,000 training samples are defined as head classes (13 categories), those with 500 to 5,000 samples are defined as medium classes (11 categories), and those with fewer than 500 samples are defined as tail classes (6 categories). 

The detailed mAP and mAUC results across these frequency-based groups are reported in Table~\ref{tab:map_auc_comparison}. In particular, we focus on the performance of tail classes to assess whether the proposed method effectively improves the recognition of rare diseases under long-tailed distribution settings.

As shown in Table~\ref{tab:map_auc_comparison} , most baseline models exhibit a significant performance degradation on tail classes when handling long-tailed distributions. For instance, although the conventional network ResNet-50 and the multimodal model CLIP achieve strong recognition performance on head classes, their mAP drops sharply to approximately 0.01 when evaluated on tail classes. This indicates that under extremely limited positive samples, the learning process of conventional feature extraction networks tends to be dominated by head classes, thereby substantially limiting their ability to learn discriminative representations for tail lesions.Meanwhile, although ConvNeXt+LDAM incorporates a long-tailed loss optimization strategy and attempts to alleviate class imbalance by strengthening decision boundaries for tail classes, its mAP on tail classes is only 0.0328. This further suggests that relying solely on post-hoc weighting constraints in the loss function is insufficient to fundamentally address the issue of missing fine-grained feature representations caused by extreme data scarcity.

Compared with the baseline methods, the proposed TRCGL-Net demonstrates stronger robustness to long-tailed distributions and significantly improves the recognition performance of tail classes. Specifically, TRCGL-Net achieves a tail-class mAP of 0.4904, which is an improvement of 0.3624 over the best-performing baseline model ConvNeXt. Meanwhile, its tail-class mAUC is also increased to 0.9229.These performance gains can be attributed to two main factors. First, at the data level, the text-guided conditional diffusion model leverages semantic priors of disease descriptions to generate high-quality synthetic samples for tail classes, thereby alleviating the issue of feature scarcity in underrepresented categories. Second, at the feature level, the Channel-Class Attention module integrates a channel enhancement mechanism together with a class-aware attention branch, which helps suppress irrelevant anatomical background noise and strengthens disease-relevant feature representations. Finally,at the label level, it further exploits label co-occurrence relationships, enabling tail classes to benefit from additional discriminative information derived from frequently co-occurring diseases.enabling tail classes to benefit from additional discriminative information from frequently co-occurring diseases. As a result, the proposed model achieves a more balanced classification performance across head, medium, and tail categories.

\begin{table}[htbp]
\centering
\caption{Performance comparison across head, medium, and tail classes in terms of mAP and mAUC}
\label{tab:map_auc_comparison}
\begin{tabular}{lcccccc}
\hline
 & \multicolumn{3}{c}{mAP} & \multicolumn{3}{c}{mAUC} \\
\cline{2-7}
Method & Head & Medium & Tail & Head & Medium & Tail \\
\hline
ConvNeXt~\cite{ref28} 
& 0.5156 & 0.2715 & 0.1280 
& 0.8882 & 0.8589 & 0.8707 \\

ResNet~\cite{ref25} 
& 0.4060 & 0.1344 & 0.0104 
& 0.8610 & 0.8176 & 0.8513 \\

EfficientNet~\cite{ref27} 
& 0.3711 & 0.0992 & 0.0076 
& 0.8319 & 0.7785 & 0.6464 \\

CLIP~\cite{ref30}  
& 0.4555 & 0.1726 & 0.0111 
& 0.8650 & 0.8186 & 0.6482 \\

ViT~\cite{ref29} 
& 0.3815 & 0.1125 & 0.0073 
& 0.8399 & 0.7883 & 0.6412 \\

ConvNeXt + LDAM~\cite{ref14} 
& 0.4903 & 0.2352 & 0.0328 
& 0.8783 & 0.8414 & 0.6898 \\

Ours 
& \textbf{0.5479} & \textbf{0.3272} & \textbf{0.4904} 
& \textbf{0.9012} & \textbf{0.8849} & \textbf{0.9229} \\
\hline
\end{tabular}
\end{table}

\subsection{Ablation Study}

To evaluate the individual contributions of the key components in TRCGL-Net, an ablation study is conducted. The complete TRCGL-Net is taken as the full model, and three ablated variants are constructed by progressively removing the Prompt-Latent Diffusion module, the Channel-Class Attention module, and the Label Relation Modeling module, respectively. All variants are trained and evaluated under exactly the same data split, input resolution, optimization strategy, batch size, and evaluation metrics to ensure a fair and consistent comparison across different configurations. In particular, no changes are made to the ConvNeXtV2 backbone or training strategy across all experiments. The experimental results are reported in Table~\ref{tab:ablation}.

\begin{table}[htbp]
\centering

\caption{Ablation study of key components}
\label{tab:ablation}
\scalebox{0.85}{
\makebox[\textwidth][c]{%
\begin{tabular}{cccccc}
\toprule
Prompt-Latent Diffusion & Channel-Class Attention & Label Relation Modeling & mAP$\uparrow$ & mAUC$\uparrow$ & F1$\uparrow$ \\
\midrule
$\times$ & $\times$ & $\times$ & 0.3671 & 0.8581 & 0.3227 \\
$\surd$ & $\times$ & $\times$ & 0.4361 & 0.8983 & 0.4192 \\
$\surd$ & $\surd$ & $\times$ & 0.4381 & 0.9000 & 0.4053 \\
$\surd$ & $\surd$ & $\surd$ & \textbf{0.4408} & \textbf{0.8989} & \textbf{0.4369} \\
\bottomrule
\end{tabular}%
}
}

\end{table}

As shown in Table~\ref{tab:ablation}, when using only ConvNeXtV2 as the feature extraction backbone for multi-label classification, the model achieves an mAP of 0.3671, indicating that although the backbone network is capable of extracting general features from chest X-ray images, it remains insufficient for long-tailed multi-label classification scenarios. Due to the significant imbalance in sample distribution across different disease categories, the training process is likely to be dominated by high-frequency classes, resulting in inadequate learning of discriminative features for low-frequency diseases and consequently limiting the overall classification performance.

With the introduction of the learnable text-guided diffusion module, the model achieves an mAP of 0.4361, representing an 18.8\% improvement over the baseline ConvNeXtV2. This result indicates that the learnable text-guided diffusion module is the primary contributor to the overall performance gains, effectively alleviating the issue of tail-class sample scarcity at the data level.

Upon further incorporating the Channel-Class Attention module, the model achieves a slight but consistent improvement to 0.4381 mAP, suggesting that joint channel-wise reweighting and category-aware attention-based feature aggregation effectively enhance tail-class discriminability.

Finally, when the Label Relation Modeling module is introduced, the full TRCGL-Net reaches the best performance of 0.4408 mAP, 0.8989 mAUC, and 0.4369 F1-score. This improvement indicates that explicitly modeling label dependencies further enhances the classifier’s ability to exploit co-occurrence relationships among diseases, leading to more robust multi-label predictions.

Overall, the learnable text-guided diffusion model, the Channel-Class Attention module, and the Label Relation Modeling module jointly improve the performance of TRCGL-Net, effectively addressing key challenges in long-tailed chest X-ray classification, including the scarcity of tail-class samples, the susceptibility of lesion representations to background interference, and the insufficient exploitation of label co-occurrence information.

\subsection{Parameter Analysis}
\subsubsection{Synthetic Sample Weighting Experiment}
The synthetic sample weighting controls the contribution ratio of tail-class samples generated by the learnable text-guided conditional diffusion model during training, thereby balancing the information proportion between synthetic and real samples. Three weighting values, 0.25, 0.5, and 0.75, are evaluated in the experiments. Table~\ref{tab:synthetic_weight} summarizes the performance metrics under different weights, including mean Average Precision (mAP), mean Area Under the Curve (mAUC), and macro-averaged F1 score.

Overall, the results show a consistent impact of different weighting settings on model performance, where mAP and mAUC achieve their best performance under a moderate weight (0.5), reaching 0.4408 and 0.8989, respectively, while F1 slightly improves to 0.4495 under a higher weight (0.75). This indicates that there exists a certain trade-off among different evaluation metrics.

Specifically, when the synthetic sample weight is low (0.25), the model underutilizes diffusion-generated samples, limiting the effectiveness of tail-class augmentation and thus degrading overall discriminative performance. When the weight is too high (0.75), the model becomes overly dependent on synthetic data during training, which may bias the learned feature distribution toward generated samples and lead to a slight decrease in mAP.

In contrast, the moderate weight (0.5) achieves a better balance between real and synthetic samples. It allows the model to fully leverage the complementary information provided by diffusion-generated tail-class samples while avoiding over-reliance on synthetic data, thereby achieving a superior trade-off between overall classification performance and training stability.

\begin{table}[htbp]
\centering
\caption{Synthetic sample weighting experiment results}
\label{tab:synthetic_weight}
\begin{tabular}{c ccc}
\hline
Synthetic Weight & mAP $\uparrow$ & mAUC $\uparrow$ & F1 $\uparrow$ \\
\hline
0.25 & 0.4173 & 0.8979 & 0.4325 \\
0.50 & \textbf{0.4408} & \textbf{0.8989} & 0.4369 \\
0.75 & 0.4324 & 0.8969 & \textbf{0.4495} \\
\hline
\end{tabular}
\end{table}

\subsubsection{Label Graph Fusion Strength Experiment}
The label graph fusion strength controls the influence of the graph convolutional branch based on label co-occurrence relationships in the final prediction, thereby regulating the propagation intensity of inter-class dependency information within the model. In this experiment, the fusion strength is set to 0.025, 0.05, and 0.1 to evaluate its impact on overall model performance.

Overall, Table~\ref{tab:label_fusion} show a consistent effect of different fusion strengths on model performance. The best performance in terms of mAP and mAUC is achieved when the fusion strength is set to 0.05, reaching 0.4408 and 0.8989, respectively, while the F1 score attains its highest value of 0.4407 at a fusion strength of 0.1, indicating a certain trade-off between mAP and F1.

Specifically, when the fusion strength is relatively low (0.025), the label graph branch contributes weakly to the propagation of inter-class dependency information, making it difficult for tail classes to fully exploit structural information from correlated high-frequency classes, thereby limiting overall discriminative performance. When the fusion strength is moderate (0.05), label co-occurrence information is effectively integrated with visual features, enhancing dependency modeling while avoiding excessive interference with visual representations, thus achieving the best performance in ranking-based metrics such as mAP.

When the fusion strength is further increased (0.1), the contribution of the label graph branch to the final prediction becomes more dominant, strengthening the influence of inter-class dependency information on decision making. This can improve local consistency in classification decisions, thereby increasing the F1 score. However, it may also introduce overly strong dependency propagation, slightly degrading mean Average Precision (mAP), indicating a partial imbalance between visual feature learning and label dependency modeling.

\begin{table}[htbp]
\centering
\caption{Label graph fusion strength experiment results}
\label{tab:label_fusion}
\begin{tabular}{c ccc}
\hline
Fusion Strength & mAP $\uparrow$ & mAUC $\uparrow$ & F1 $\uparrow$ \\
\hline
0.025 & 0.4279 & 0.8962 & 0.4248 \\
0.050 & \textbf{0.4408} & \textbf{0.8989} & 0.4369 \\
0.100 & 0.4211 & 0.8987 & \textbf{0.4407} \\
\hline
\end{tabular}
\end{table}

\subsection{Case Study}
To further evaluate the interpretability and discriminative reliability of TRCGL-Net in long-tailed multi-label chest X-ray classification, a visualization analysis is conducted on both class-specific response regions and prediction results. Fig.~\ref{fig:gradcam} presents the visualization of class-discriminative regions across different chest disease samples, where the red bounding boxes on the original chest radiographs indicate manually annotated suspicious abnormal regions, and the heatmaps below represent the spatial response intensity of the model when predicting the corresponding classes.Fig.~\ref{fig:prediction} further illustrates the multi-label prediction results and corresponding confidence scores for different test samples, which are used to analyze the model’s discriminative capability in complex multi-label scenarios.

\subsubsection{Visualization of Class Decision Regions}
\begin{figure}[htbp]
\centering
    \centering
    \includegraphics[width=1\textwidth]{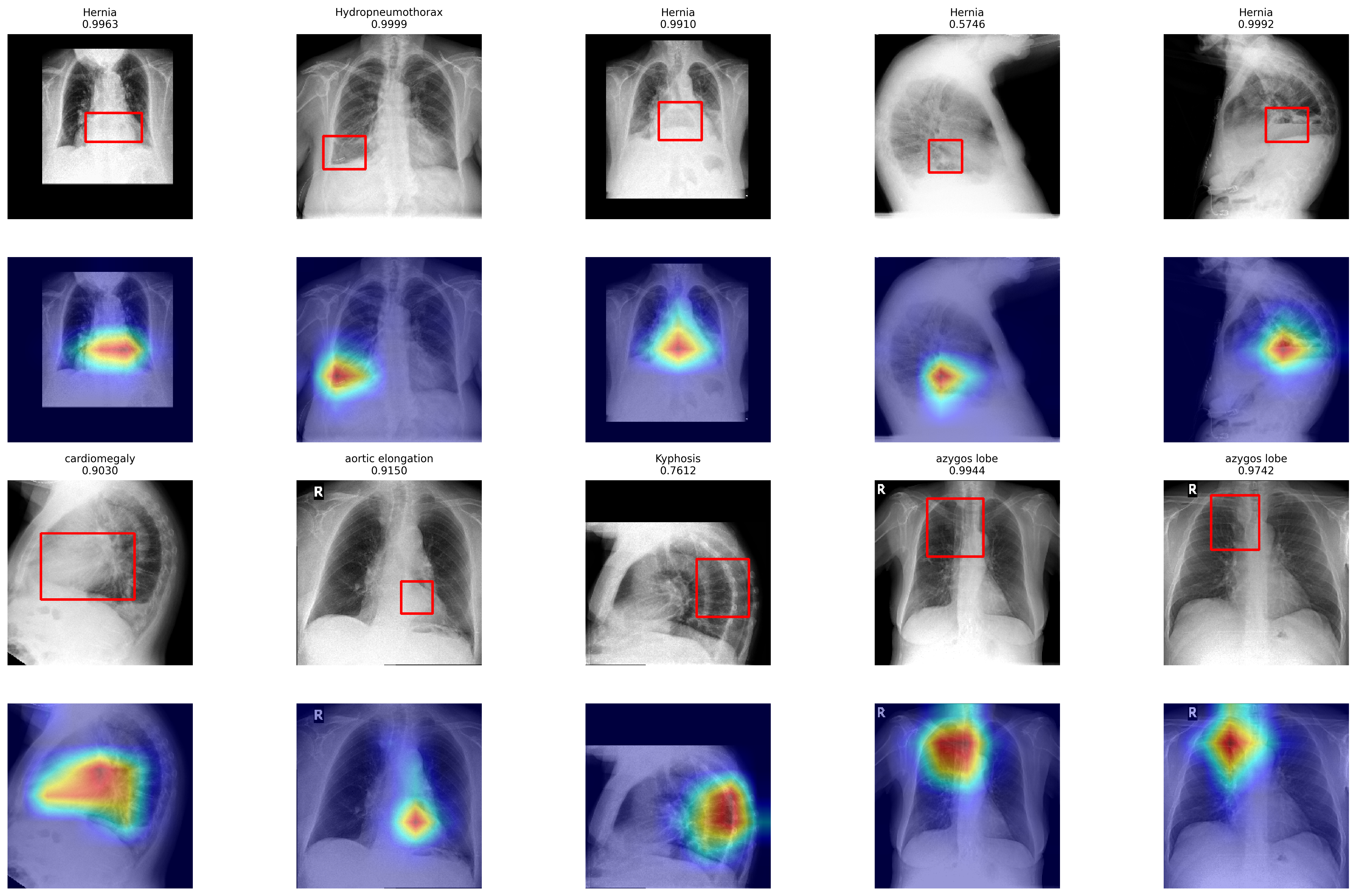}
    \caption{Visualization of Class Decision Regions in TRCGL-Net}
    \label{fig:gradcam}
\end{figure}

As shown in Fig.~\ref{fig:gradcam}, TRCGL-Net can accurately focus on lesion-relevant local regions across multiple types of thoracic abnormalities. For samples such as cardiomegaly, aortic elongation, pulmonary parenchymal abnormalities, and kyphosis, the regions with high activation are predominantly concentrated around the cardiac silhouette boundaries, mediastinal structures, localized abnormal regions in the lung fields, or vertebral structures. These findings are consistent with the anatomical regions typically relied upon for disease discrimination in chest X-ray imaging.This indicates that the model is not only capable of leveraging global image semantics for classification, but also effectively highlights class-specific discriminative local regions through a class-aware attention aggregation mechanism. Consequently, it reduces interference from irrelevant background structures such as ribs, clavicles, and soft tissue shadows, thereby improving the robustness of model predictions.

\subsubsection{Multi-label prediction results and corresponding confidence scores for different test samples}
\begin{figure}[htbp]
\centering
    \centering
    \includegraphics[width=1\textwidth]{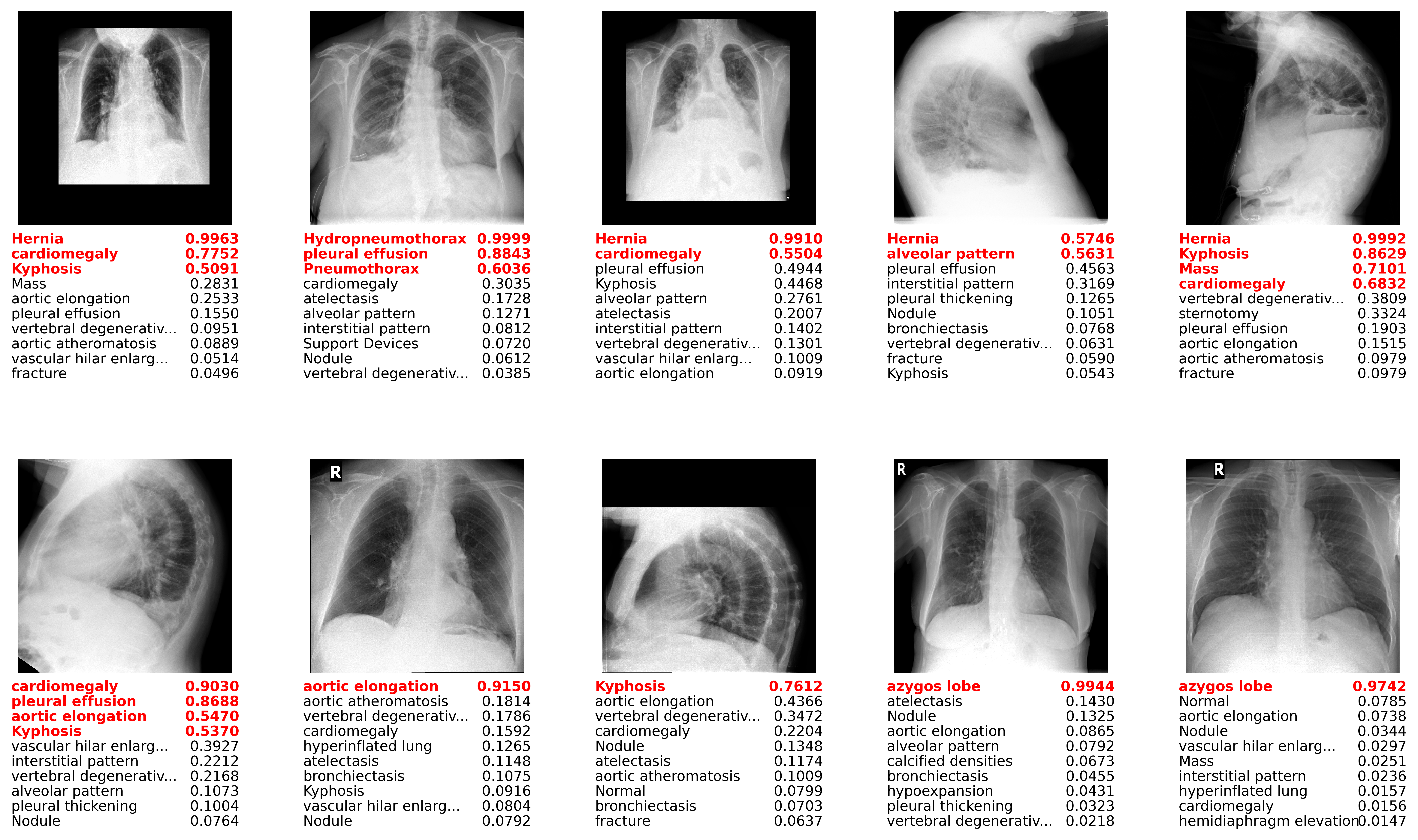}
    \caption{Visualization of Multi-Label Prediction Confidence in TRCGL-Net}
    \label{fig:prediction}
\end{figure}

Fig.~\ref{fig:prediction} illustrates the multi-label prediction confidence results on the test samples. It can be observed that TRCGL-Net is capable of simultaneously assigning multiple potential abnormality labels within a single chest radiograph and allocating corresponding prediction probabilities for different categories. Furthermore, for certain samples, the model is able to predict accompanying labels that are associated with the primary lesion, such as cardiomegaly-related abnormalities, aortic alterations, pulmonary texture changes, and structural abnormalities. This indicates that modeling label co-occurrence relationships helps the model capture latent dependencies between diseases in complex chest X-ray images.
\FloatBarrier
\section{Conclusion}

To address the challenges of long-tailed multi-label chest X-ray classification, including severe sample imbalance in tail classes, subtle lesion characteristics, and insufficient exploitation of label co-occurrence information, this study proposes the TRCGL-Net framework. The proposed method integrates a learnable text-guided conditional diffusion model with feature enhancement and label-relation modeling strategies, enabling comprehensive optimization at the data, feature, and label-relation levels for tail-class learning.

At the data level, the conditional diffusion model generates synthetic samples for tail classes under disease-semantic guidance, effectively alleviating data scarcity and improving the diversity of rare disease patterns. At the feature level, channel-wise feature recalibration and class-aware spatial attention mechanisms are introduced to enhance discriminative representations while suppressing interference from irrelevant anatomical structures. At the label level, a graph-based label co-occurrence modeling strategy is employed to capture inter-class dependencies, enabling tail classes to benefit from informative relationships with high-frequency categories.

Ablation studies and visualization results demonstrate that each component of TRCGL-Net contributes positively to overall classification performance and tail-class recognition capability. Moreover, the model can adaptively focus on key localized regions corresponding to different disease categories within the same chest radiograph, thereby improving both discriminative reliability and interpretability.

Although the proposed method achieves promising performance in long-tailed multi-label chest X-ray classification, several limitations remain. First, the learnable text-guided conditional diffusion model relies on disease names and radiology reports as semantic conditions. Its semantic representation capacity may still be limited in more complex clinical scenarios, such as multi-lesion descriptions or fine-grained subtype differentiation. Second, the label co-occurrence graph is constructed based on training-set statistics, which effectively captures disease dependencies under the current data distribution. However, such dependencies may vary across institutions or shifting data distributions, indicating that more adaptive label-relation modeling strategies are still required.

Future work will focus on two main directions. First, more structured medical knowledge guidance will be explored, incorporating anatomical priors or report-level clinical semantics into the diffusion process to further improve the medical consistency and expressiveness of synthetic tail-class samples. Second, dynamic and data-adaptive label relation modeling methods will be investigated, enabling the label co-occurrence structure to evolve with data distribution shifts, thereby enhancing robustness and generalization in cross-domain settings.

\section*{Acknowledgements}
This work was supported by the Fundamental Research Funds for the Central Universities of South-Central Minzu University (Grant Number: CZQ24015).

\section*{Conflict of Interest}
The authors declare no competing interests.

\section*{Data Availability}
The pretrained ConvNeXtV2 weights are publicly available at: \url{https://huggingface.co/hieuphamha/cxrlt2026-task1-convnextv2}

Code are publicly available at:   \url{https://github.com/November-1113/TRCGL-Net}
%
%
%
%
\bibliographystyle{unsrt}
\bibliography{ref}

@inproceedings{dong2026isbi,
  title={Overview of the cxr-lt 2026 challenge: Multi-center long-tailed and zero shot chest x-ray classification},
  author={Dong, Hexin and Lin, Yi and Zhou, Pengyu and Feng, Xuan Zhong and Legasto, Alan Clint and Lin, Mingquan and Chen, Hao and Yang, Yuzhe and Shih, George and Peng, Yifan},
  booktitle={2026 IEEE 23rd International Symposium on Biomedical Imaging (ISBI)},
  pages={1--4},
  year={2026},
  organization={IEEE}
}

@article{cxrlt2024survey,
  title={Towards long-tailed, multi-label disease classification from chest X-ray: Overview of the CXR-LT challenge},
  author={Holste, Gregory and Zhou, Yiliang and Wang, Song and Jaiswal, Ajay and Lin, Mingquan and Zhuge, Sherry and Yang, Yuzhe and Kim, Dongkyun and Nguyen-Mau, Trong-Hieu and Tran, Minh-Triet and others},
  journal={Medical Image Analysis},
  volume={97},
  pages={103224},
  year={2024},
  publisher={Elsevier}
}

@inproceedings{cxrlt2024noisy,
  title={Long-tailed multi-label classification with noisy label of thoracic diseases from chest X-ray},
  author={Lai, Haoran and Yao, Qingsong and He, Zhiyang and Tao, Xiaodong and Zhou, S Kevin},
  booktitle={2024 IEEE International Symposium on Biomedical Imaging (ISBI)},
  pages={1--5},
  year={2024},
  organization={IEEE}
}

@article{ref4,
  title={Multi-label chest x-ray image classification with single positive labels},
  author={Xiao, Jiayin and Li, Si and Lin, Tongxu and Zhu, Jian and Yuan, Xiaochen and Feng, David Dagan and Sheng, Bin},
  journal={IEEE transactions on medical imaging},
  volume={43},
  number={12},
  pages={4404--4418},
  year={2024},
  publisher={IEEE}
}

@inproceedings{ref5,
  title={Multi-Label Disease Detection in Chest X-Ray Imaging Using a Fine-Tuned ConvNeXtV2 with a Customized Classifier},
  author={Xiong, Kangzhe and Tu, Yuyun and Rao, Xinping and Zou, Xiang and Du, Yingkui},
  booktitle={Informatics},
  volume={12},
  number={3},
  pages={80},
  year={2025},
  organization={MDPI}
}

@article{ref6,
  title={Chexnet: Radiologist-level pneumonia detection on chest x-rays with deep learning. arXiv},
  author={Rajpurkar, Pranav and Irvin, Jeremy and Zhu, Kaylie and Yang, Brandon and Mehta, Hershel and Duan, Tony and Ding, Daisy and Bagul, Aarti and Langlotz, Curtis and Shpanskaya, Katie and others},
  journal={arXiv preprint arXiv:1711.05225},
  volume={10},
  year={2017}
}

@article{ref7,
  title={An optimized transformer model for efficient detection of thoracic diseases in chest X-rays with multi-scale feature fusion},
  author={Yu, Shasha and Zhou, Peng},
  journal={Plos one},
  volume={20},
  number={5},
  pages={e0323239},
  year={2025},
  publisher={Public Library of Science San Francisco, CA USA}
}

@article{ref8,
  title={IEViT: An enhanced vision transformer architecture for chest X-ray image classification},
  author={Okolo, Gabriel Iluebe and Katsigiannis, Stamos and Ramzan, Naeem},
  journal={Computer Methods and Programs in Biomedicine},
  volume={226},
  pages={107141},
  year={2022},
  publisher={Elsevier}
}

@article{ref9,
  title={Label correlation transformer for automated chest X-ray diagnosis with reliable interpretability},
  author={Sun, Zexuan and Qu, Linhao and Luo, Jiazheng and Song, Zhijian and Wang, Manning},
  journal={La radiologia medica},
  volume={128},
  number={6},
  pages={726--733},
  year={2023},
  publisher={Springer}
}

@article{ref10,
  title={Comparison of deep learning approaches for multi-label chest X-ray classification},
  author={Baltruschat, Ivo M and Nickisch, Hannes and Grass, Michael and Knopp, Tobias and Saalbach, Axel},
  journal={Scientific reports},
  volume={9},
  number={1},
  pages={6381},
  year={2019},
  publisher={Nature Publishing Group UK London}
}

@article{ref11,
  title={Bag of tricks for long-tailed multi-label classification on chest x-rays},
  author={Hong, Feng and Dai, Tianjie and Yao, Jiangchao and Zhang, Ya and Wang, Yanfeng},
  journal={arXiv preprint arXiv:2308.08853},
  year={2023}
}

@inproceedings{ref12,
  title={Enhancing multi-label long-tailed classification on chest x-rays through ML-GCN augmentation},
  author={Seo, HyeRyeong and Lee, MinHyuk and Cheong, WooJin and Yoon, HyeKyung and Kim, SoHyung and Kang, MyungJoo},
  booktitle={Proceedings of the IEEE/CVF international conference on computer vision},
  pages={2747--2756},
  year={2023}
}

@article{ref13,
  title={Padchest: A large chest x-ray image dataset with multi-label annotated reports},
  author={Bustos, Aurelia and Pertusa, Antonio and Salinas, Jose-Maria and De La Iglesia-Vaya, Maria},
  journal={Medical image analysis},
  volume={66},
  pages={101797},
  year={2020},
  publisher={Elsevier}
}

@article{ref14,
  title={Loss Design and Architecture Selection for Long-Tailed Multi-Label Chest X-Ray Classification},
  author={Sulake, Nikhileswara Rao},
  journal={arXiv preprint arXiv:2603.02294},
  year={2026}
}

@article{ref15,
  title={LTCXNet: Advancing Chest X-Ray Analysis with Solutions for Long-Tailed Multi-Label Classification and Fairness Challenges},
  author={Huang, Chin-Wei and Shen, Mu-Yi and Shih, Kuan-Chang and Lin, Shih-Chih and Chen, Chi-Yu and Kuo, Po-Chih},
  journal={arXiv preprint arXiv:2411.10746},
  year={2024}
}

@article{ref16,
  title={Evaluation of deep convolutional generative adversarial networks for data augmentation of chest x-ray images},
  author={Kora Venu, Sagar and Ravula, Sridhar},
  journal={Future Internet},
  volume={13},
  number={1},
  pages={8},
  year={2020},
  publisher={MDPI}
}

@article{ref17,
  title={DualAttNet: Synergistic fusion of image-level and fine-grained disease attention for multi-label lesion detection in chest X-rays},
  author={Xu, Qing and Duan, Wenting},
  journal={Computers in Biology and Medicine},
  volume={168},
  pages={107742},
  year={2024},
  publisher={Elsevier}
}

@article{ref18,
  title={Automated thorax disease diagnosis using multi-branch residual attention network},
  author={Li, Dongfang and Huo, Hua and Jiao, Shupei and Sun, Xiaowei and Chen, Shuya},
  journal={Scientific Reports},
  volume={14},
  number={1},
  pages={11865},
  year={2024},
  publisher={Nature Publishing Group UK London}
}

@article{ref19,
  title={CLARiTy: A Vision Transformer for Multi-Label Classification and Weakly-Supervised Localization of Chest X-ray Pathologies},
  author={Statheros, John M and Wang, Hairong and Klein, Richard},
  journal={arXiv preprint arXiv:2512.16700},
  year={2025}
}

@article{ref20,
  title={Label co-occurrence learning with graph convolutional networks for multi-label chest x-ray image classification},
  author={Chen, Bingzhi and Li, Jinxing and Lu, Guangming and Yu, Hongbing and Zhang, David},
  journal={IEEE journal of biomedical and health informatics},
  volume={24},
  number={8},
  pages={2292--2302},
  year={2020},
  publisher={IEEE}
}

@article{ref21,
  title={Modeling global and local label correlation with graph convolutional networks for multi-label chest X-ray image classification},
  author={Li, Lanting and Cao, Peng and Yang, Jinzhu and Zaiane, Osmar R},
  journal={Medical \& Biological Engineering \& Computing},
  volume={60},
  number={9},
  pages={2567--2588},
  year={2022},
  publisher={Springer}
}

@inproceedings{ref22,
  title={Label Semantic Improvement with Graph Convolutional Networks for Multi-Label Chest X-Ray Image Classification},
  author={Cai, Dachuan and Lu, Huijuan and Chai, Zhuijun and Wang, Renfeng and Zhu, Wenjie and Yao, Yudong},
  booktitle={2023 13th International Conference on Information Technology in Medicine and Education (ITME)},
  pages={711--717},
  year={2023},
  organization={IEEE}
}

@article{ref23,
  title={Graph guided multiscale cross attention for multilabel chest X ray classification},
  author={Shi, Guokun and Wang, Zijian and Shi, Yucheng and Pan, Jingwen and Sun, Liping and Fang, Fang and Jin, Li},
  journal={Scientific Reports},
  year={2026},
  publisher={Nature Publishing Group UK London}
}

@inproceedings{ref24,
  title={Learning transferable visual models from natural language supervision},
  author={Radford, Alec and Kim, Jong Wook and Hallacy, Chris and Ramesh, Aditya and Goh, Gabriel and Agarwal, Sandhini and Sastry, Girish and Askell, Amanda and Mishkin, Pamela and Clark, Jack and others},
  booktitle={International conference on machine learning},
  pages={8748--8763},
  year={2021},
  organization={PmLR}
}

@article{ref25,
  title={Deep residual learning for image recognition: A survey},
  author={Shafiq, Muhammad and Gu, Zhaoquan},
  journal={Applied sciences},
  volume={12},
  number={18},
  pages={8972},
  year={2022},
  publisher={MDPI}
}

@inproceedings{ref27,
  title={Efficientnet: Rethinking model scaling for convolutional neural networks},
  author={Tan, Mingxing and Le, Quoc},
  booktitle={International conference on machine learning},
  pages={6105--6114},
  year={2019},
  organization={PMLR}
}

@inproceedings{ref28,
  title={A convnet for the 2020s},
  author={Liu, Zhuang and Mao, Hanzi and Wu, Chao-Yuan and Feichtenhofer, Christoph and Darrell, Trevor and Xie, Saining},
  booktitle={Proceedings of the IEEE/CVF conference on computer vision and pattern recognition},
  pages={11976--11986},
  year={2022}
}

@article{ref29,
  title={An image is worth 16x16 words: Transformers for image recognition at scale},
  author={Dosovitskiy, Alexey and Beyer, Lucas and Kolesnikov, Alexander and Weissenborn, Dirk and Zhai, Xiaohua and Unterthiner, Thomas and Dehghani, Mostafa and Minderer, Matthias and Heigold, Georg and Gelly, Sylvain and others},
  journal={arXiv preprint arXiv:2010.11929},
  year={2020}
}

@inproceedings{ref30,
  title={Learning transferable visual models from natural language supervision},
  author={Radford, Alec and Kim, Jong Wook and Hallacy, Chris and Ramesh, Aditya and Goh, Gabriel and Agarwal, Sandhini and Sastry, Girish and Askell, Amanda and Mishkin, Pamela and Clark, Jack and others},
  booktitle={International conference on machine learning},
  pages={8748--8763},
  year={2021},
  organization={PmLR}
}

@inproceedings{ref31,
  title={Handling supervision scarcity in chest x-ray classification: Long-tailed and zero-shot learning},
  author={Pham, Ha-Hieu and Nguyen, Hai-Dang and Nguyen, Thanh-Huy and Xu, Min and Bagci, Ulas and Le, Trung-Nghia and Pham, Huy-Hieu},
  booktitle={2026 IEEE 23rd International Symposium on Biomedical Imaging (ISBI)},
  pages={1--4},
  year={2026},
  organization={IEEE}
}

\end{document}